\documentclass{IEEEtran}
\usepackage{subfigure}
\usepackage{moreverb}
\usepackage{epsfig}
\usepackage{amsmath,amssymb,amsthm,mathrsfs,amsfonts,dsfont}
\usepackage{adjustbox,lipsum}
\usepackage{algorithm,algorithmic}
\usepackage{amsfonts}
\usepackage{epsfig}
\usepackage{amssymb}
\usepackage{amsmath}
\usepackage{amsthm}
\usepackage{subfigure}
\usepackage{multirow}
\usepackage{rotating}
\usepackage{graphicx}
\usepackage{tabularx}
\usepackage{array}
\usepackage{anyfontsize}
\usepackage{color,soul}
\usepackage{graphicx,dblfloatfix}
\usepackage{epstopdf}
\usepackage{blindtext}
\usepackage{amsthm,amssymb,amsmath,bm}
\usepackage{subfigure}
\usepackage{amsfonts}
\usepackage{epsfig}
\usepackage{amssymb}
\usepackage{amsmath}
\usepackage{cite}
\usepackage{graphicx}
\usepackage{fancyhdr}
\usepackage{subfigure}
\usepackage[subfigure]{tocloft}
\usepackage[font={small}]{caption}
\usepackage{subfigure}
\usepackage{tabularx}
\usepackage{cite}

\begin{document}
\title{Artificial Intelligence as a Services (AI-aaS) on Software-Defined Infrastructure } 

\author{
	\IEEEauthorblockN{ Saeedeh Parsaeefard\IEEEauthorrefmark{1}, Mahsa Derakhshani\IEEEauthorrefmark{2}}\\
	\IEEEauthorblockA{\IEEEauthorrefmark{1} Communication Technologies \& Department, ITRC, Tehran, Iran} \\
	\IEEEauthorblockA{\IEEEauthorrefmark{2}Wolfson School of Mechanical, Electrical \& Manufacturing Engineering,  Loughborough University, UK}
	\vspace{-12mm}
}

\author{Saeedeh Parsaeefard, Iman Tabrizian, Alberto Leon-Garcia\\Electrical and  Computer Engineering Depratement, University of Toronto\\ saeideh.fard@utoronto.ca, iman.tabrizian@mail.utoronto.ca,
	alberto.leongarcia@utoronto.ca
	
}
\maketitle
\begin{abstract}
    This paper investigates a paradigm for offering artificial intelligence as a service (AI-aaS) on software-defined infrastructures (SDIs). The increasing complexity of networking and computing infrastructures is already driving the introduction of automation in networking and cloud computing management systems. Here we consider how these automation mechanisms can be leveraged to offer AI-aaS. Use cases for AI-aaS are easily found in addressing smart applications in sectors such as transportation, manufacturing, energy, water, air quality, and emissions. We propose an architectural scheme based on SDIs where each AI-aaS application is comprised of a \textit{monitoring, analysis, policy, execution plus knowledge} (MAPE-K) loop (MKL). Each application is composed as one or more specific service chains embedded in SDI, some of which will include a Machine Learning (ML) pipeline. Our model includes a new training plane and an AI-aaS plane to deal with the model-development and operational phases of AI applications. We also consider the role of an ML/MKL sandbox in ensuring coherency and consistency in the operation of multiple parallel MKL loops. We present experimental measurement results for three AI-aaS applications deployed on the SAVI testbed: 1. Compressing monitored data in SDI using autoencoders; 2. Traffic monitoring to allocate CPUs’ resources to VNFs; and 3. Highway segment classification in smart transportation. 
\end{abstract}
\begin{IEEEkeywords}
   Artificial intelligence as a service (AI-aaS), cognitive network management, MAPE-K loop, kubernetes, Kube-flow, MEC, micro service management.
\end{IEEEkeywords}
\vspace{-.55cm}
\IEEEpeerreviewmaketitle
\section{Introduction}

Artificial intelligence (AI) aims to provide autonomous/cognitive system behavior using diverse approaches from big data analytics and machine learning (ML). The vast range of applications sectors and verticals include many instances where AI is applied in a \textit{monitoring, analysis, policy, execution plus knowledge} (MAPE-K) loop to manage the internal operation of a system as well as its interactions with other systems in an autonomous manner \cite{bookmapek}. In MAPE-K loop (MKL), the collection and aggregation of data, its analysis by analytics and ML engines, and the decision-making can be viewed as a chain of functions executed over an integrated communication and computation infrastructure such as software defined infrastructure (SDI) \cite{sdi}. We consider offering MKLs as a new service over SDIs which we refer to as AI as a service (AI-aaS). We refer to each graph of MKL functions as an MKL-chain. 

There are many AI-aaS use cases ranging from control system for robots in factories, to traffic monitoring in smart cities, and prominently autonomous network management in 5G \cite{itutreport}. Each AI-aaS application has an associated set of quality and performance requirements that need to be translated into quality of service (QoS) requirements that must be met by resources in SDI. Clearly, SDIs should be highly agile, flexible and dynamic to serve these diverse use cases \cite{itutreport}. In particular, there has been a surge in activities to explore how to implement MKLs for network automation \cite{boutaba2018,kdn,eni,itutreport,3gpp2,atis2018,acumos,onap}.

The notion of SDI was introduced in \cite{sdi} as a multi-tier cloud of virtualized networking and computing resources managed by an integrated management system  orchestrating end-to-end (E2E) resources to support distributed applications. The development of software defined networking (SDN) and network function virtualization (NFV), edge mobile computing (MEC), and virtualization of wireless access have led to one prominent architecture for SDI in 5G, e.g., \cite{8444568}. However, alternative architectures for SDI continue to emerge stimulated by advance in computing, specifically modular and microservices-based structures, containerization and associated monitoring and orchestration, i.e., KubeFlow \cite{bibentry2019apr}, data-center based on FPGA virtualization, novel networking capabilities in Linux kernel, and recent advances in AI engine software and systems, e.g., scikit-learn, TensorFlow, PyTorch, ML pipelines \cite{itutreport}, Acumos \cite{acumos}, and ONAP \cite{onap}.

These new capabilities along with autonomous MKLs in networking and its diverse use cases motivates revisiting SDI architecture based on MKLs' requirement. In this paper, we aim to address these practical issues based on a new emerging standard draft for ML pipelines \cite{itutreport}. We investigate AI-aaS use cases based on their QoS for each step of MKL. We propose a nominal SDI architecture to handle MKL chains. From SDI perspective, we categorize AI-aaS use cases in two main groups: network management applications (NAL) which are related to internal autonomous network management and control loops; and over the top (OTT) applications which are served by SDI as slices. To control potential conflicts of parallel running MKLs, we introduce new entities in the network management plane. To train and re-train MKLs in off-line or on-line modes, we introduce local sandboxes for each use cases and we introduce a new sandbox to investigate the mutual effects of parallel MKL-chains to manage the stability. Using Kubeflow \cite{bibentry2019apr}, we develop three use cases for AI-aaS. The first two use cases are related to NAL where autoencoders are applied for data compression, and the effect of the traffic on the required resources of each VNF is investigated, and then, the required resource per each VNFs is predicted. For the OTT AI applications, with CVST outputs from smart city project in the greater Toronto area \cite{cvst2019may}, we classify the highways based on the speed of vehicles during 24 hours.

\begin{figure*}
	\begin{center}
		\includegraphics[width=6.3 in]{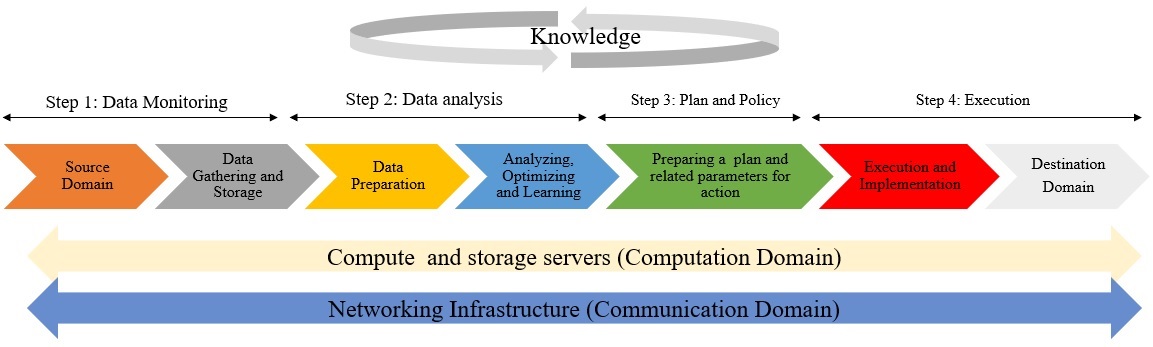}
		\caption{MAPE-K loop (MKL) for AI assisted management and learning in systems (AI-aaS)}
		\label{MAPEK}
	\end{center}
\end{figure*}


This paper is organized as follows. Section II reviews the MKL and AI-aaS's use cases. Section III studies modified SDIs to support AI-aaS. In Section IV,  implementation use cases are introduces; followed by conclusions in Section V.

\section{MAPE-K Loop and Use Case Presentation of AI-aaS}

AI is a set of functions to realize learning behavior via classification, regression, reasoning, planning, knowledge representation, search or any other types of functions based on data analysis and perceive information from the data \cite{atis2018}. AI involves diverse disciplines and provides capabilities to handle problems with high computational complexity, to deal with unknown environments and extract new features, and to assist in MKLs in any system. MKLs are the main focus of this paper. One illustration of MKLs is presented in Fig. \ref{MAPEK} where each step can be described as follows: 

\textbf{Step 1:} \underline{M}onitoring an environment via sensor devices, measurement tools, and collecting data within a suitable time window. In this step, data gathering and storage can be provided by SDIs. The source domain is a place that data is gathered from. e.g., highways in smart city via sensors or cameras.   

\textbf{Step 2:} \underline{A}nalysis of data based on the different functions in an AI context (AI engines) to support a solution. AI engines need to be developed to reach a solution for a given use case, which may encompass the followings: 
	\begin{itemize}
		\item Data preparations such as filtering, de-noising, normalization, de-normalizations; 
		\item Knowledge creation engines e.g., classifications, segmentation, association, regression anomaly detection, prediction, inference engine or semantic reasoner; 
		\item Decision support or decision making engines to yield the desired solution, optimization tools and reinforcement learning. AI engines can be based on supervised learning, unsupervised learning, and reinforcement learning \cite{8553661}.   
	\end{itemize}

\textbf{Step 3:} \underline{P}lanning and policy about results of Step 2 including translation of the results of Step 2 to parameters understandable by the system, e.g., adjusting the traffic lights in smart city, and transmit power allocation of users in 5G. Scheduling of set of actions based on results of Step 2 can be handled here. 
	
\textbf{Step 4:} \underline{E}xecuting the action which can be deployed autonomously or by human intervention. The destination domain includes a set of nodes that should execute the actions, e.g., a set of users in 5G which should change their transmit power or a set of robots which should change their states. 

These steps can be repeated until the learning process converges (in the machine learning process). In another case, a training phase can be deployed in offline manner to evaluate the outputs of the AI process. Each step of MAPE-K can be removed in AI-aaS to cover more use cases. Each step of MAPE-K can be removed in AI-aaS to cover more use cases.

From above, evidently, one MKL is equivalent to a chain or a graph of functions (MKL-chain) which should be run in a specific order over an SDI \cite{itutreport}. MKL-chains should be places in an SDI based on their requirements. From an SDI perspective, classifying the use cases into classes with similar performance requirements would be ideal, this is not trivial for AI-aaS use cases due to their diverse sets of requirements. To deal with this challenge, we propose that for each block in Fig. \ref{MAPEK}, the following items are determined: 1) Communication domain parameters, e.g., the minimum required bandwidth, throughput, reliability and a coverage area should be determined; 2) Computation domain parameters, e.g., minimum and maximum amount of data, and speed and reliability for both computation and storage servers. Other factors such as security and privacy also can be considered for each use case. Afterwards, the appropriate parameters per each domain and step are defined, e.g., minimum required bandwidth for Step 1 to collect information from the field. 
\begin{figure}
	\begin{center}
		\includegraphics[width=3.7 in]{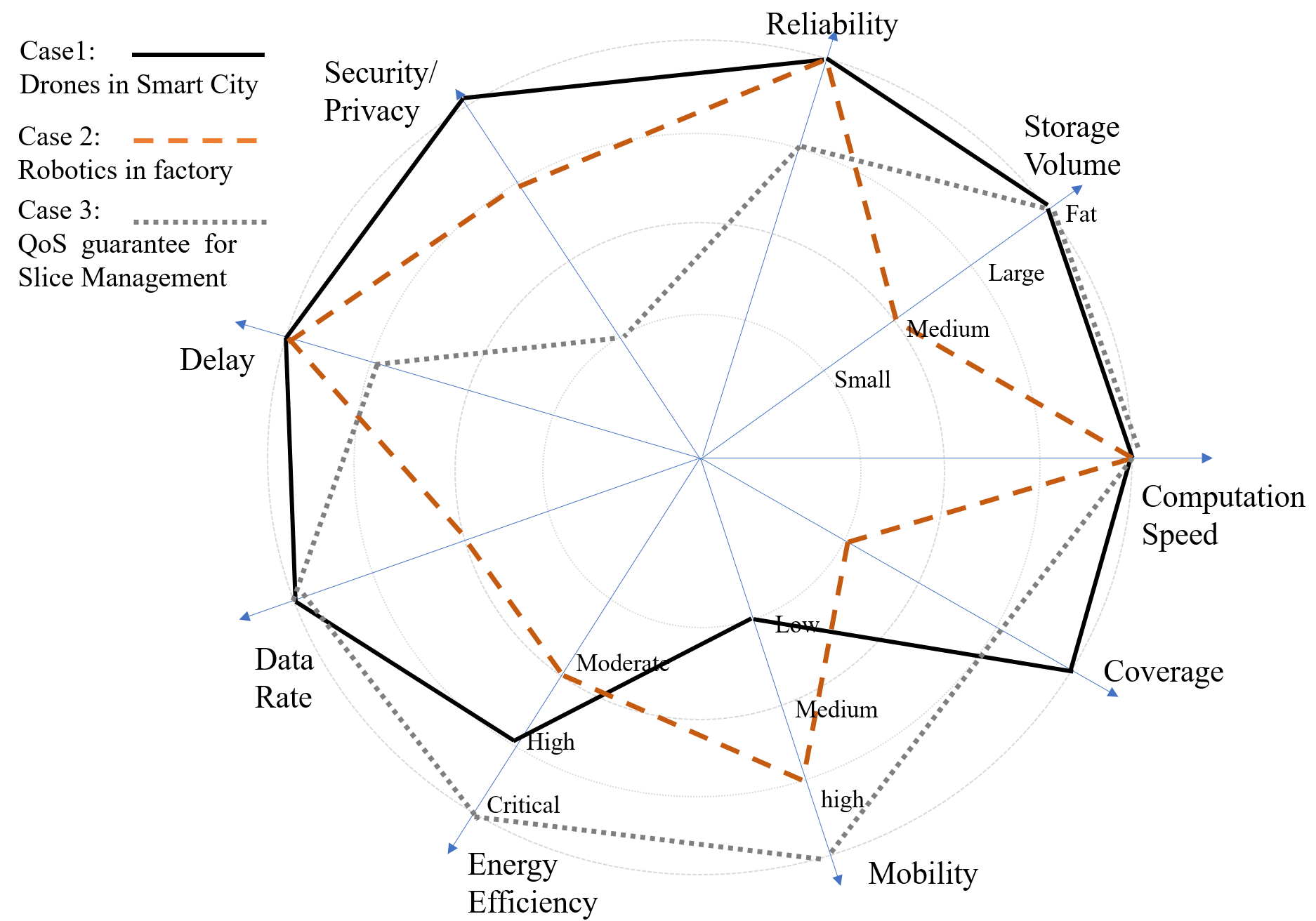}
		\caption{Spider diagrams for AI-aaS use cases based on required E2E key features, e.g., delay, data rate, energy efficiency, mobility, coverage, computation speed, storage volume, and reliability. }
		\label{Spidercharts}
	\end{center}
\end{figure}

For example in Fig. \ref{Spidercharts}, the spider diagrams are presented for three use cases: 1) Case 1: traffic monitoring in smart city; 2) Case2: mobility control of robots in factory; and, 3) Case 3: QoS management for network slicing. For case 1, MKL can be defined as a monitoring and capturing images from the highways by a set of cameras or sensors, sending captured images to one computing center via a provided SDI. Then, on-line analysis is developed based on different image processing and anomaly detection in AI context, e.g., via neural networks. The results are passed to control center in a  smart city to deploy specific procedures. 
The communication domain for this scenario should provide high data rates for large number of sensors or cameras in smart city area while considering the energy efficiency. Energy efficiency depends on deployment scenarios of sensors and cameras. The volume of stored images may be high, and the AI engines to prepare data, analysis and abnormal detection should be run vary quickly. SDI should be highly reliable and secure. For Case 2, when an object is to design a MKL-chain to handle the mobility of robots and their state control in a factory, the first step is to collect information from a filed in the factory. The coverage area of this case is smaller than that of Case 1. The states of robots and their locations can be sent by highly reliable and secure links but with limited amount of throughput. The amount of processing and storage is also not considerable compared to those in Case 1, but E2E delay is critical in order to make a decision about the next state of robots. For Case 3, we deal with QoS management for slices of one network operator based on a users' traffic pattern of each slice. Here, all the states of entities in the network and end users' traffic parameters should be collected and analyzed. Then, based on a number of AI engines, the QoS of each user of each slice should be estimated \cite{3gpp1}. Allocated resources of each slice can be determined based on its users' traffic and QoS prediction, and the states of network's nodes. Here, the source and destination domains are inside the network, and for computation domain, high capacity storage units and high speed computation servers should be available for Step 2. The action also should be implemented in a real time manner to guarantee the users' QoS. These use cases demonstrate the similarities in MKLs of AI-aaS use cases while their diverse QoSs for each step which leading to different design of SDIs.
\begin{figure*}
	\begin{center}
		\includegraphics[width=6.9 in]{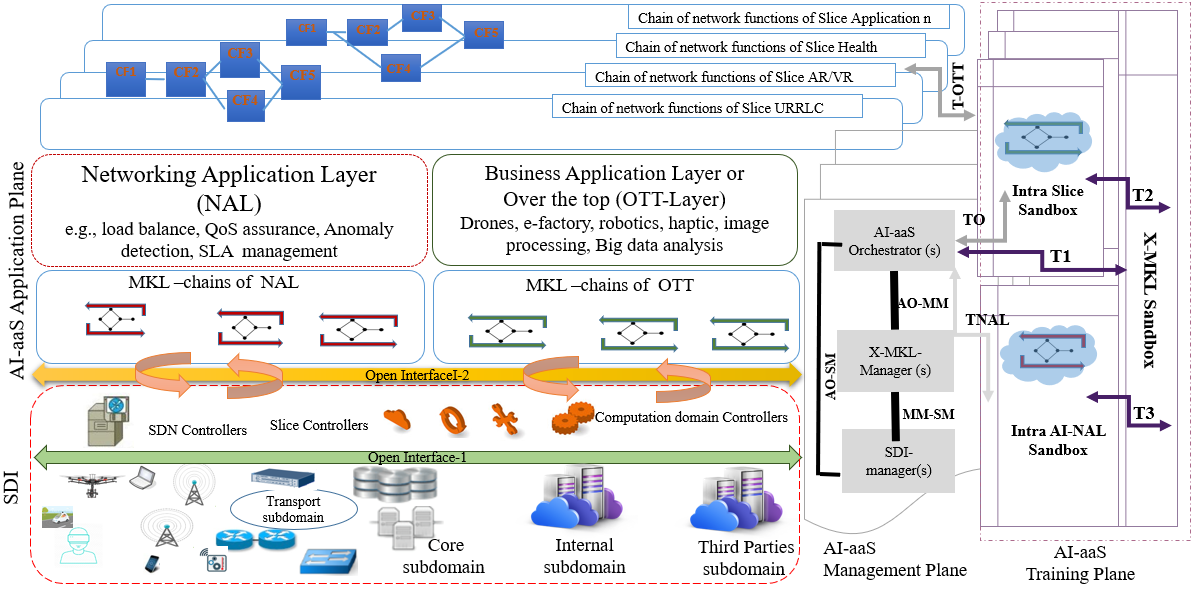}
		\caption{Multi-sandbox SDI architecture including management elements to control the interference between MKLs}
		\label{AIAAS}
	\end{center}
\end{figure*}    


\section{SDI based Architecture to Serve AI-aaS }
Evidently, to support AI-aaS in an SDI architecture, the communication and computation domains should be more integrated and have a more flexible design. Also, the network management and application planes should be modified to support MKLs and MKL-chains. Fig. \ref{AIAAS} depicts a possible architectures, explained as follows: 
  
\subsubsection{SDI plane} This is a multi-tier cloud based on programmable and softwarized integrated communication and computation infrastructure, e.g., \cite{sdi}. In SDI, communication domain has radio access network (RAN), transport and core subdomains, equipped with software defined radios (SDRs) and E2E SDN and NFV. SDN provides logically centralized controllers to handle the network functionalities and application program interfaces to adjust the network processes dynamically \cite{sdncomprehensivesurvey}. SDI is empowered via NFV where each network function can be run virtually over any available servers via a concept of virtual network function (VNF). In SDI, SDN and NFV can be leveraged to apply MKL in autonomous network management \cite{kdn}. SDR provides programmability in the access nodes of RAN in which  via function splitting in RAN \cite{8479363} and providing a cloud close to the end users, e.g., mobile edge computing (MEC), a more flexible architecture for RAN is available to serve AI-aaS use cases. In the computation domain, based on microservices structures over containers, any AI engines can be deployed in SDI (from edge via MEC to core), or in the third parties clouds.

Hence, in SDI, each application request can be considered as a specific graph of VNFs refereed to an application chain. Via orchestration and management concepts, i.e., MANO, this chain is placed in a SDI based on its QoS requirements, e.g., for real-time applications. Therefore, SDIs have a unified view of any application and provides a substrate to offer AI-aaS use cases. The graph will be placed in SDI resources based on its QoS requirements, e.g., for highly delay sensitive applications, the graph of application is placed in a proximity of end-users and the MEC or cloud in edge handles the required computations. Based on this unified view, similarly, AI-aaS applications are considered as MKL-chains implemented in SDIs where both physical/ virtual computing and communication platforms are prepared to preserve requirements of AI-aaS applications. The SDI plane provides coexistence of third parties, license and unlicensed frequency bands, and heterogeneous access nodes. SDI also includes control elements e.g., SDN controllers where between forwarding elements and control entities, there exists open interface 1 which has two upward and downward directions to pass data of the network to controllers and transferring the action to the source for MKL-chains of network applications. Here, SDI includes all westbound and eastbound interfaces between communication and computation controllers.




\subsubsection{ MKL-chain and AI-aaS Application Plane} which are responsible to represent each AI-aaS use case based on SDIs functions and QoS requirements. MKL-chain are graphs providing orders and connections between functions from Steps 1 to 4 in Fig. \ref{MAPEK}. Source domain and destination domain are places where data should be gathered and actions should be executed, respectively. For network applications in AI-aaS, source and destination are inside of the communication domain, while for other applications, the source and destination is in the field of use cases, i.e., in a factory or in a smart city region. For Step 3, a catalog function can be introduced to translate the output of AI engines to an understandable parameter by the network or OTT application \cite{onap}. The functions for each MKL-chain, depending on their communication or computation needs, can be mapped to physical or virtual network functions in the communication domain, or specific container, storage functionalities or microservices in the computation domain. Each use case can request SDI to provide a service to run this MKL-chain. Depending on the SLA and cost, all steps of each use case can be handled by the SDI of an operator or with the collaboration of third parties. Via this view, the main process to perform any AI-aaS use case is to embed its MKL-chain into SDI based on the required QoS and the SDI's state. The \textit{AI-aaS application plane} takes care of all MKL-chains for slices as well as MKL-chains for network applications (e.g., Case 3 in Fig \ref{Spidercharts}) and OTT applications (e.g., Case 1 and Case 2 in Fig \ref{Spidercharts}). This plane is also responsible to provide an E2E QoS intent per each MKL-chain and report it to a management plane. Between an AI-aaS application and SDI, there exists an open access Interface 2 which is responsible to provide interactions between these two planes.

\subsubsection{AI-aaS Management plane} This contains three levels of management components and it is responsible for on-boarding, monitoring and terminating any new requests from AI-aaS use cases and traditional network applications. At the lowest level of management, it includes SDI manager(s) to tune and adjust all physical and virtual multi-tier computing, communication and cloud resources to preserve all AI-aaS application requirements \cite{sdi}. It also can include VNF managers to handle VNF's functionalities. Cross (inter and intra) MKL manager (X-MKL manager) performs configuration and KLM life-cycle management (e.g., instantiation, update, query, scaling, termination) on different computing and computation domains. The other important task of this entity is to verify the  accuracy, coherency and consistency of the output of MKLs-chains as well as their interference to each other. The AI-aaS orchestrator inherits all the responsibility of the orchestrator in SDI \cite{sdi}, MANO, slice management, and traditional network management. However, it should be equipped with new procedures to handle the priorities of MKLs, stability analysis of network and other systems when more than one parallel MKLs are run simultaneously with the help of X-MKL and training plane, explained next. 
	
\subsubsection{AI-aaS Training Plane} This provides an environment for training, retraining and examining any feature in MKLs in offline and online manners \cite{itutreport}. However, to consider different features and structures of AI-NAL and AI-OTT applications, we propose a multi-sandbox structure in which there exist three categories of sandboxes: 1) Intra slice sandbox which is responsible to provide an environment to have full emulation/simulation/pilot environment for each slice and OTT applications; 2) Intra AI-NAL Sandbox which takes care of emulation of each MKL-chain of network applications in the network; and 3) Cross MKL-chain sandbox (X-MKL Sandbox) which is responsible to simulate and emulate the interaction and effects of outputs of MKLs on each others and sending reports to the orchestartor about conflicting situations, and also determining their priorities and orders for parallel MKLs. The X-MKL sandbox helps the management plane to remove any chance of instability in the network proposed by NAL or OTT MKLs. In order to emulate an environment, the training plane communicates with the management plane to receive all the state information of the SDI. This type of interactions is handled via T-NAL interface. T-OTT is an interface between slices and OTT applications and slices to pass information about the training phase to the application part. T1, T2 and T3 are responsible to pass information between orchestrator, intra sandboxes of OTT and AI-NAL sandboxes, respectively. Other interfaces are also provided in the Fig. \ref{AIAAS}. X-MKL sandbox is also associated to preserve the privacy and security of information between different applications.  

\begin{figure*}
	\centering
	\includegraphics[width=4.3 in]{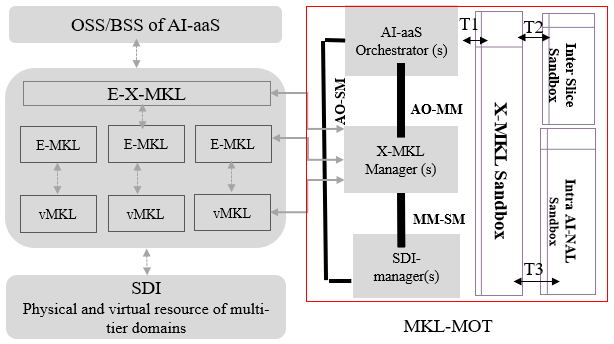}
	\caption{The MKL management, orchestration and training (MKL-MOT) architectural framework.}
	\label{fig:MANO-AIaaS}
\end{figure*}

Based on the above planes and E2E virtual structure of SDI, the MKLs can be considered in a virtual manner, i.e., virtual MKL (vMKL), which also can be  managed and trained via MKL- management, orchestration and training (MKL-MOT) architectural frameworks depicted in Fig. \ref{fig:MANO-AIaaS}. Here, for each vMKL, there is an element management of MKL (E-MKL) which is responsible for the network management functions FCAPS (Fault,
Configuration, Accounting, Performance, and Security) of a running vMKL. All of these are connected to Element of management of cross interaction between MKLs (E-X-MKL). This element is responsible for preserving priority, coherency of outcome of vMKLs, and preventing their interference. This type of information is provided by X-MKL sandbox which passes this information to the AI-aaS orchestrator. AI-aaS orchestrator wtogether with the X-MKL manager and E-X-MKL handle these issues. The related interfaces between different elements are depicted in Fig. \ref{fig:MANO-AIaaS}.    

\begin{figure*}
	\centering
	\includegraphics[width=5.3 in]{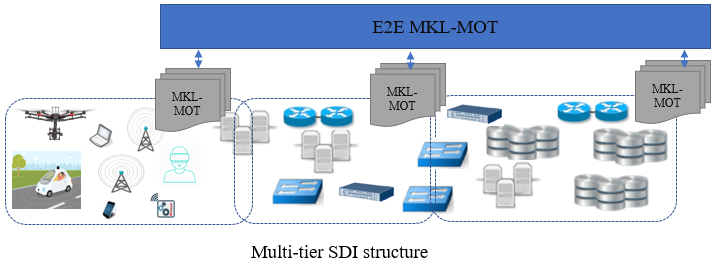}
	\caption{Hierarchical and recursive structure of MKL-MOT.}
	\label{fig:End-toend}
\end{figure*}

For the multi-tier and multi-domain infrastructure different domains may have their own time scales and their own applications. The recursive structure of MKL-MOT can be applied as in Fig. \ref{fig:End-toend}, in which interactions of MKL-MOTs of different tiers can be handled by a hierarchically-based E2E  MKL-MOT. Therefore, a diverse time scales of different parameters in the network can be handled more efficiently. For instance, in the NAL use cases, the transmit power of access nodes needs to be more frequently adjusted compared to the VNF size of firewalls in the core domain. Or in  robots in a factory, the states of robots are updated more frequently than the sensor parameters in a factory.  

\section{Standardization and Industrial Activities}
There has been a surge of standardization and industrial activities to bring AI in networking and to offer AI services from communication and computation organizations, e.g., \cite{cognet,selfnet}. 
In Table \ref{111}, we compare them with AI-aaS in terms of their applications, stability assurance of network for parallel MKLs, and sandbox's features with the most recent activities. 
 \begin{table}	\centering
 	\caption{{Comparing Recent Activities with AI-aaS}}
 	\begin{tabular}{llll}
 		\hline 
 		& Unified  & Interaction  & Sandbox  \\
 		&view & assurance  &    \\
 	\hline	\hline
 		AI-aaS	& $\surd$  & $\surd$ &Networking $\&$ OTT   \\
 		ML-pipeline\cite{itutreport}	& $\surd$  & $\times$ & Networking $\&$ OTT  \\
 		ENI	\cite{eni} & Networking & $\times$ & $\times$  \\
 		NWDAF\cite{3gpp1}	& Networking & $\times$ &$\times$   \\
 		ONAP\cite{onap}	& Networking  & $\times$  & $\times$  \\
 		ACUMOS \cite{acumos}& Computing & $\times$ &  OTT  \\
 		\hline \hline
 	\end{tabular}	\label{111}
 \end{table}   

\textit{Network data analytic function (NWDAF)} is introduced by 3GPP \cite{3gpp1} to collect and analysis data of one network operator to offer data driven slice management  \cite{3gpp2}. \textit{Experiential networked intelligence (ENI)} is introduced by ITU-T to bring a closed-loop of AI mechanisms in network management \cite{eni} where there are two new top-down and bottom-up interfaces in ENI which are mainly responsible to translate the output of AI engine layer into the actionable network element parameters, and to sense and report the network information to AI-engine layer, respectively. 

\textit{ ML Pipe Line} is introduced by ITU-T where a set of logical entities should be chained in order to provide the AI functionalities and closed loop control \cite{itutreport}. This context is similar to our MKK-chain while the latter is based on MAPE-K context. \textit{ACUMOS} is an open source AI machine learning platform to develop any AI based applications in "\textit{containerized microservices}" and modular manner \cite{acumos}.  \textit{Open network automation platform (ONAP)} is a microservice platform to bring a full network management automation in 5G for both physical and virtual network functions \cite{onap}. In ONAP, there is a closed loop automation management platform (CLAMP) to provide any AI-based operation within networks. 


\section{Deployment Scenarios of IaaS Via Kubeflow }

\begin{figure*}
	\centering
	\includegraphics[width=4.3in]{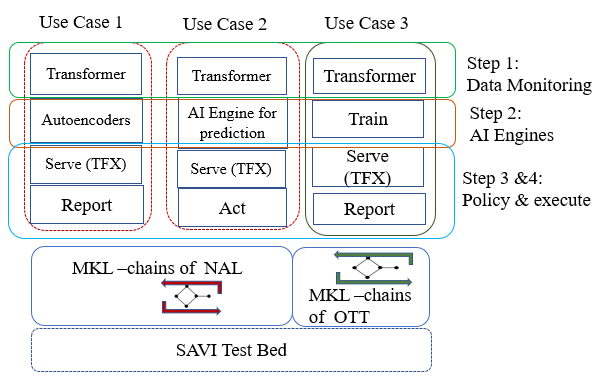}
	\caption{Three use cases of AI-aaS on SAVI Testbed \cite{kang2013may} and their MKL-chains based on Kubeflow}
	\label{fig:topology}
\end{figure*}

We use Kubeflow \cite{bibentry2019apr} and Prometheus to deploy AI-aaS for both networking and OTT applications over the SAVI infrastructure \cite{savitestbed}. The SAVI testbed is a Canada-wide multi-tier heterogeneous testbed with edges in Victoria, Waterloo, Calgary, and Carlton and headquarter in Toronto. Kubeflow is an open-source Google project that facilitates deployment of E2E machine learning pipelines.
The scheme in our implementation setup is depicted in Fig. \ref{fig:topology} where an MKL-chain for each application is shown. 

\subsubsection{Use Case 1: Compressing Monitored Data via Autoenconders in Networking Applications }
\begin{figure}
	\centering
	\includegraphics[width=2.6 in]{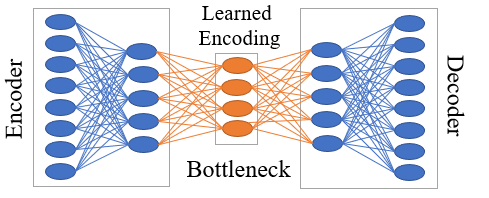}
	\caption{General Architecture of Autoencoder }
	\label{fig:autoencoder}
\end{figure}

One of the main issues for a cognitive network management is how to compress diverse types of data in more efficient manner and reduce a huge volume of data in networking. Here, we show how autoencoders in an MKL-chain can handle this issue efficiently. The MKL-chain of this application includes monitoring phase of data of each function, nodes, or VN in network. Then, the data is passed to the AI engines responsible for compressing data (Step 2 of MAPE-K). Afterwards, the compressed data is stored or passed to other MKL-chain or components in the networks. For the compression phase via AI engines, different approaches can be applied. Designing an appropriate compressing algorithm is not trivial. We compress monitored data for network management via autoencoders. General architecture of autoencoders is shown in Fig. \ref{fig:autoencoder}. In autoencoders, the number of neurons in each layer decreases as the depth of the neural network increases until the "bottleneck layer". Bottleneck layer contains the encoded data. The encoded data can be used for reconstruction of the original data. In our implementation, the number of features in the bottleneck equals to 75. From the bottleneck layer, the number of neurons in each layer increases. The layers from the input layer until the bottleneck layer is called "encoder". The neural network consisting of layers from the bottleneck layer to the output layer is called "decoder". We apply the autoencoder introduced in \cite{rumelhart1985learning} with the architecture described in Table \ref{tab:autoencoder}.

\begin{table}[]
	\centering \caption{Autoencoder Architecture of Setup}
	\begin{tabular}{llll}
		\hline
		Layer Number & Activation Function & \# of Inputs & \# of Outputs  \\ \hline \hline
		1 & Elu & 111  &  90 \\ \hline
		2 & Elu & 90  &  85 \\ \hline
		3 & linear & 85  &  75 \\ \hline
		4 & Elu & 85  &  90 \\ \hline
		5 & Sigmoid & 90  &  111 \\ \hline
	\end{tabular}
		\label{tab:autoencoder}
\end{table}

\begin{figure}
	\centering
	\includegraphics[width=0.4\textwidth]{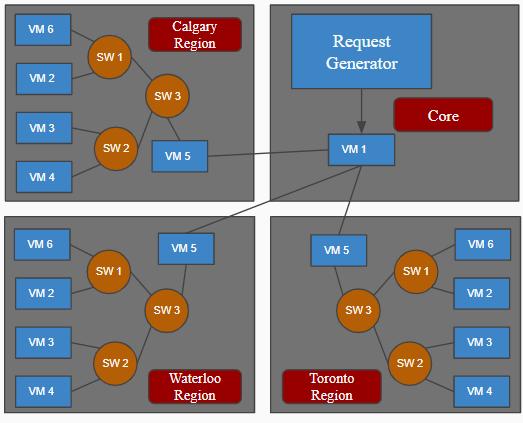}
	\caption{ Topology for Compressed Data in Monitoring}
	\label{fig:experiment-topology}
\end{figure}

\begin{figure}
	\centering
	\includegraphics[width=0.45\textwidth]{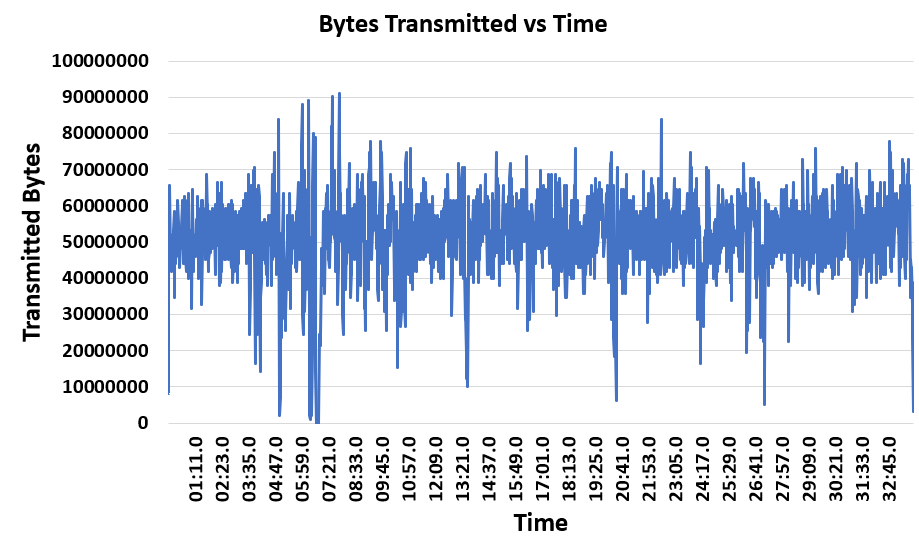}
	\caption{Traffic Pattern of HAProxy load balancer for 30 minutes}
	\label{fig:traffic-pattern}
\end{figure}

We apply the aforementioned autoencoder for compressing the time-series data collected from an emulated network on SAVI \cite{kang2013may} and depicted in Fig. \ref{fig:experiment-topology}. It includes 16 VMs (blue boxes) and 9 VMs acting as switches (orange circle). Each VM has Open vSwitch and a preinstalled monitoring tool. For each region, there is a Prometheus server, responsible for pulling metrics from the VMs and a Ryu SDN controller to setup the flow rules. There is an HAProxy load balancer (VM 1) installed in the "Core", which passes the requests to the web servers, i.e., VM 6 in each region. It is installed in Toronto, Waterloo and Calgary regions to emulate realistic network delays. The HTTP traffic applied to the HAProxy (VM 1 in Core) is depicted in Fig. \ref{fig:traffic-pattern}. Then, we build our data-set by collecting metrics from Prometheus deployed in each region. The list of some of metrics is provided in Table \ref{tab:metrics}. This data-set includes 111 types of data and we aim to reduce the number of metrics required for storage by around 30\%. To train the autoencoder, we split the data-set into validation and training sub-sets. Our data-sets are available in https://www.nal.utoronto.ca/autoencoder-ai-aa-s which is based on 30 minutes snap shot of data. We first train the autoencoder with the training set. We evaluate its performance using the validation set. We use "Mean Square Error" as a cost function and "elu" as an activation function for the neurons. The MKL-chain is implemented in Kubeflow. When the training phase is completed, the autoencoder is served using TFX \cite{46484} to compress monitored data in real time.

\begin{figure}
	\centering
	\includegraphics[width=0.5\textwidth]{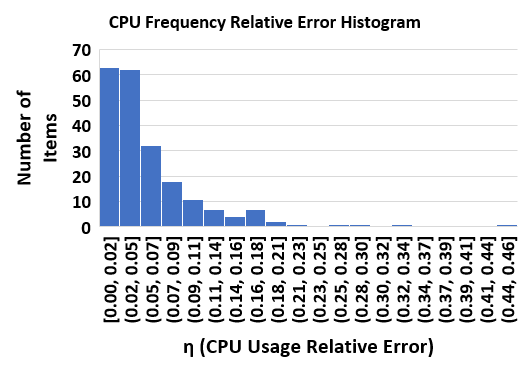}
	\caption{Error Distribution of CPU Usage Compression}
	\label{fig:compare-cpu}
\end{figure}

An important issue in compressing data is about the reconstruction of data. To evaluate the autoencoder decompression phase, in Fig. \ref{fig:compare-cpu}, we investigate the error distribution of reconstructing "CPU usage" data for VM 6 in region Waterloo. In Fig.\ref{fig:compare-cpu}, horizontal axis is a relative error of the desired parameter, i.e., $\eta_{\text{CPU usage}}=\frac{\text{CPU usage real}-\text{CPU usage reconstruct}}{\text{CPU usage real}}$. From Fig. \ref{fig:compare-cpu}, more than $85\%$ of "CPU usage" data is reconstructed with less than $10\%$ error. This result is promising for the potential of autoencoders in networking applications. In https://www.nal.utoronto.ca/autoencoder-ai-aa-s, we present more results of the autoencoder for our setup.  


\begin{table}[]
    \centering  \caption{List of Some Metrics}
    \begin{tabular}{ll}
\hline
         Metric Name & Description  \\ \hline \hline
         node\_cpu & CPU Percentage \\ \hline
         node\_network\_receive & Node Network Bytes Received \\ \hline
         node\_network\_transmit & Node Network Bytes Transmitted \\ \hline
         container\_memory\_usage & Container Memory Usage  \\ \hline
         http\_request\_size & HTTP Request Size \\ \hline
         http\_request\_duration & HTTP Request Duration \\ \hline
         container\_cpu\_system & Container CPU System Usage \\ \hline
         container\_network\_receive & Container Network Bytes Received \\ \hline
         container\_network\_transmit & Container Network Bytes Transmitted \\ \hline
    \end{tabular}
  
    \label{tab:metrics}
\end{table}


\subsubsection{Use Case 2: Adaptive Resource Allocation for VNFs} 
\begin{figure}
	\centering
	\includegraphics[width=0.32\textwidth]{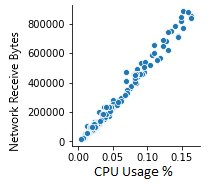}
	\caption{Traffic vs CPU}
	\label{fig:traffic-vs-cpu}
\end{figure}

Here, we aim to provide more efficient resource allocation for VNFs based on traffic prediction. Therefore, we need to predict traffic as well as the relation between required resources and network traffic. Here, the emulated network is similar to use case 1. In each region, there is a firewall application deployed in VM 4 and a web server running in VM 6 and all traffic generated in VM 5 should be passed through Snort firewall installed in VM 4. After filtering malicious traffic, it is steered through VM 6 in each region (its original destination). SVOP (Simple VNF Orchestration Platform) \cite{svop} takes care of the deployment of Snort and installation of flow rules. To allocate a CPU to the Snort efficiently, we should predict the required CPU based on an amount of traffic passed through Snort. Here, we have four steps of MKL-chain as presented in Fig. \ref{fig:topology} where the data gathered by Prometheus is stored in "Object Storage" in "Transformer" phase. The output of "Transformer" is the data-set for training the ML model. We need two sets of AI engines: 1) a set predicting the amount of CPU required for each VNF based on network traffic; 2) a second set predicting the traffic for a specific period of time, i.e., the next 10 minutes for our setup. Then, in the "Serve" phase, we apply TFX \cite{46484} to serve two sets of AI engines over HTTP. For the first set of AI engines, we collect two hours data using Prometheus. Then, we apply linear regression which reveals that there is a linear relation between network traffic and CPU usage of Snort (as depicted in Fig. \ref{fig:traffic-vs-cpu}) with MSE 0.000003. Consequently, if we can predict a traffic of the network, we can linearly adjust the CPU for Snort to reach a better resource utilization. For the prediction of network traffic, we deploy Long short-term memory (LSTM) which is an artificial recurrent neural network (RNN) architecture

\subsubsection{Use Case 3: Classifying Highways in Smart Cities Based on Speed Patterns of Vehicles} Due to page limitation, we only provide a brief setup of MKL-chain of this use case in Fig. \ref{fig:topology} and more information is provided in \cite{savitestbed}. We will provide more detail information about these types of use cases of AI-aaS in SAVI test bed in near future.

\section{Conclusion }
We investigate the concept of artificial intelligence as a service (AI-aaS) where SDI serves both AI-based network management and over the top applications in a similar manner. This unified view opens a new avenue for business opportunities in future networking. We propose a nominal network architecture for AI-aaS, define new elements in the network management plane to guarantee the stability of MAP-K loops introduced by AI, and control any their outcomes' conflicts. Using Kubernetes and Kubeflow, we deploy three applications for data compressing, resource management of VNFs, and traffic classifications in smart city. AI-aaS and its related concepts are underdeveloped and there is a surge of need to study different aspects of this context, e.g., 1) placing MKL-chains in SDIs while satisfying their QoS, 2) Composition and re-composition of MKL-chain; 3) Security and privacy quarantines of passing information in different sandboxes; 4) QoS guarantee for real time and near real time MKL-chain; and 5) Distributed versus centralized MKL controllers.

\bibliographystyle{IEEEtran}

\bibliography{IEEEabrv,AIaaS}

\end{document}